\documentclass{llncs}

\usepackage{courier}
\usepackage{soul}
\usepackage{graphicx}
\usepackage{amsmath}
\usepackage{amssymb}
\usepackage{xspace}
\usepackage{float}
\usepackage{pgf}
\usepackage{booktabs}
\usepackage[bookmarks=false,colorlinks=true,urlcolor=blue,citecolor=blue]{hyperref}
\usepackage{subfig}
\usepackage{tikz}
\usetikzlibrary{shapes,arrows,shadows}

\usepackage{listings}
\newcommand{\ra}[1]{\renewcommand{\arraystretch}{#1}}
\newcommand{\tool}{\texttt{SAT Heritage}\xspace}
\newcommand{\docker}{\texttt{Docker}\xspace}
\newcommand{\zenodo}{\texttt{Zenodo}\xspace}
\newcommand{\guix}{\texttt{Guix}\xspace}
\newcommand{\github}{\texttt{GitHub}\xspace}

\begin{document}
\title{SAT Heritage: a community-driven effort for archiving, building and running more than thousand SAT solvers}

\author{Gilles Audemard\inst{1}  \and Lo{\"\i}c Paulev\'e\inst{2} \and  Laurent Simon\inst{2}}
\institute{
  CRIL, Artois University, France, email: audemard@cril.fr
  \and Univ. Bordeaux, Bordeaux INP, CNRS, LaBRI, UMR5800, F-33400 Talence, France, email: \{loic.pauleve,lsimon\}@labri.fr
}

\maketitle

\begin{abstract}

SAT research has a long history of source code and binary releases, thanks to competitions organized every year.
However, since every cycle of competitions has its own set of rules and an adhoc way of publishing source code and
binaries, compiling or even running any solver may be harder than what it seems.
Moreover, there has been more than a thousand solvers published so far, some of them released in the early 90's. If the
SAT community wants to archive and be able to keep track of all the solvers that made its history, it urgently needs
to deploy an important effort.

We propose to initiate a community-driven effort to archive and to allow easy compilation and running of all SAT solvers
that have been released so far.  We rely on the best tools for archiving and building binaries (thanks to Docker,
GitHub and Zenodo) and provide a consistent and easy way for this.
Thanks to our tool, building (or running) a solver from its source (or from its binary) can be done in one line.

\end{abstract}

\label{sec:intro}
\section{Introduction}

As Donald Knuth wrote in~\cite{knuth}, ``The story of satisfiability is the tale of a triumph of software engineering''.
In this success story of computer science, the availability of SAT solvers source code have been crucial. Archiving and
maintaining this important amount of knowledge may be as important as archiving the scientific papers that made this
domain. The release of the source code of \texttt{MiniSat}~\cite{Minisat} had, for instance, a dramatic impact on the field.
However, nothing has yet been done to ensure that source code and recipes to build SAT solvers will be archived in the
best possible way. This is a recent but important concern in the more broadly field of computer science.  The Software
Heritage~\cite{heritage} initiative is, for instance, a recent and strong initiative to handle this.
In the domain of SAT solvers, however, collecting and archiving may not be sufficient: we must embed the recipe to build
the code and to run it in the most efficient way. As input format for SAT solvers remains the same since more than 25
years~\cite{dimacs}, it is always possible to compare the performances of all existing solvers, given a suitable way of
compiling and running them. At that time, some code was using EGCS, a fork of GCC 2.8 including more features. Facebook
and Google didn't exist and Linux machines were running with kernels 1.X. Solvers were distributed with source code to
be compiled on Intel or SPARC computers. Fortunately enough, binaries for Intel 386 machines distributed at that time
are still executable on recent computers, given the availability of compatible libraries.

Collecting and distributing SAT solvers source code is, luckily, not new. SAT competitions, organized since the beginning
of the 21st century, have almost always forced the publication of the source code of submitted solvers. If source code
was not distributed, binaries were often available.  However, since the first competitions, the landscape of computer
science has changed a lot. New technologies like \docker~\cite{docker} are now available, changing the way tools are
distributed.

We propose in this work to structure and bootstrap a collective effort to maintain a comprehensive and user-friendly
library of all the solvers that shaped the SAT world. We build our tool, called \tool, on top of other recent tools,
typically developed for archiving and distributing source code and applications, like \docker \cite{docker},
\github~\cite{github}, \guix~\cite{guix}, \zenodo~\cite{zenodo2}.  The community is invited to
contribute by archiving, from now on, all the solvers used in competitions (and papers). We also expect authors of
previous solvers to contribute by adding informations about their solvers or special command lines not especially used
during competitive events. Our tool allows, for instance, to add a DOI (thanks to \zenodo) to the exact version
of any solver used in a paper, allowing simple but powerful references to be used.

In summary, the goals of our open-source tool are to:
\begin{itemize}
	\item Collect and archive all SAT solvers, binaries and sources,

	\item Easily retrieve a \docker image with the binary of any solver, directly from the \docker Hub, or, when source
		code is available, by locally building the image from the source code of the solver,

	\item Allow to easily run any SAT solver that have ever been available (typically in the last 30 years), by a one
		line call (consistent over all solvers),

	\item Open an convenient solution for reproducibility (binaries, source code and receipt to build binaries are
		archived in a consistent way), thanks to strong connection with tools like \guix and
        \zenodo.

\end{itemize}

\section{History of SAT solvers releases and publications}

The first SAT competitions happened in the 90's~\cite{competition92,competition96}. Their goals were multiple:
collect and compare SAT solvers performances in the fairest possible way, collect and distribute benchmarks, and
also take a snapshot of the performances reached so far.
Table~\ref{tab:number} reports the number of SAT solvers that took part in the different competitions. We counted more
than a thousand solvers, but even counting them was not an easy task: one source code can hide a number of subversions
(with distinct parameters) and distinct tracks, and some information were only partially available.

\begin{table}
      \centering\ra{1.1}

  \begin{tabular}{@{} r  c    c  c p{1cm}  r  c    c  c@{}}
    \toprule
Date & \#Solvers & Collection & Type && Date & \#Solvers & Collection & Type \\
\midrule
    $\leq$2000 & 24 & Satex & s / b & & 2011 & 104 & Contest    (2) & s / b \\
2002 & 27 & Contest (1) & b&&2012 & 65 & Challenge & -\\

2003 & 33 & Contest (1) & b&&2013 & 140 & Contest (3) & s(*) / b(*)\\

2004 & 63 & Contest (1) & b&&2014 & 150 & Contest (3) & s(*) / b(*) \\

2005 & 47 & Contest (1) & b&&2015 & 31 & Race (2) & - \\

2006 & 16 & Race (1) & -&& 2016 & 32 & Contest (4) & s / b \\

2007 & 31 & Contest (2) & s / b&&2017 & 71 & Contest (4) & s / b \\

2008 & 19 & Race (1) & - &&2018 & 66 & Contest (4) & s / b \\

2009 & 64 & Contest (2) & s / b&& 2019 & 55 & Race (3) & s / b\\ 

2010 & 20 & Race (1) & -   && Total & 1058 & & \\
    \bottomrule
\end{tabular}
\caption{\label{tab:number}Number of solvers to the different competitions.
  Note that some solvers may be counted twice or more (some solvers did not change from year to
  the next or have been included in a competition as reference). (*) binaries and sources are available, but by
  navigating individually to each solver result. Different numbers indicate different organizers and different way of
  distributing results, source code (s) and binaries (b).}
\end{table}

Following the ideas of these first competitions organized in the 90's, and thanks to the development of the web, the
\texttt{satex}~\cite{satex} website published solvers and benchmarks gathered by the website maintainer.
\texttt{satex} was running SAT solvers on only one personal computer. Some solvers were modified to comply with the
input/output of the \texttt{satex} framework (like a normalized exit code value). It
was a personal initiative, made possible by the relatively few solvers available (all solvers of the initial \texttt{satex} are available in our tool).

During the first cycle of competitions (numbered 1 in table~\ref{tab:number})~\cite{competition2002}, submitters had to
compile a static binary of their solver (to prevent library dependencies) via remote access to the same computer. To
ensure the deployment of their solver, this computer had the exact same Linux version as the one deployed on the cluster
used to run the contest.  Some solvers were coming from industry, which explains why no open source code was mandatory:
the priority was to draw the most accurate picture of solvers performances.  However, it was quickly decided
(competitions numbered 2 in the above table) that it was even more important to require submitters to open their code.
Binaries were then allowed to enter the competition, but only in the demonstration category (no prizes).  More recently,
thanks to the \texttt{starexec} environment~\cite{starexec}, compilation of solvers was somehow normalized (an image of
a virtual Linux machine on which the code would be built and run was distributed).
With each cycle of competition or race, came its own set of rules with an \textit{ad hoc} way of publishing source code and
binaries, with a non uniform way of providing details on which parameters to use. For example, since 2016, solvers must
provide a certificate for unsatisfiable instances~\cite{drup,drat}. One has thus to go through all the solvers
to find the correct parameters for running them without proof logging.

Thus, despite the increasing importance of software archiving~\cite{heritage}, the way SAT solvers are
distributed had not really changed in the last 25 years. It is still mainly done via personal websites, or SAT
competitions and races websites, each cycle of events defining its own rules for this. As a result, it is often unclear
how to recover any SAT solver (same code, same arguments) used in many papers, old or recent. It is even more
questionable whether, despite the importance of SAT solvers source code, we are able to correctly archive and maintain
them.


\section{SAT Heritage \docker images}

The \tool project provides a centralized repository of instructions to build and execute the SAT solvers involved in
competitions since the early ages of SAT.  To that aim, it relies on \docker images which are self-contained Linux-based
environments to execute binaries.
\docker allows to explicitly mention all the packages needed to compile the source code and to build a temporary image
(the ``builder'') for compiling the solver.
Then, the compiled solver is embedded in another, lighter, image which contains only the
libraries required to execute it.
So, each version of each collected solver is made available in a dedicated \docker image. Thanks to the layer structure
of images, all solvers sharing the same environment will share the major part of the image content, thus substantially
saving disk space. At the end, the \docker image will not be much heavier than the binary of the solver.

\docker images can be executed on usual operating systems.
On Linux, \docker offers the same performance as native binaries: only filesystem and network
operations have a slight overhead due to the isolation~\cite{docker-perf}, which is not of concern
for SAT solvers.
On other systems, the images are executed within a virtual machine, adding a noticeable
performance overhead, although considerably reduced on recent hardware \cite{docker-perf}.

\subsection{Architecture}

The instructions to build and run the collected solvers are hosted publicly on \github%
\cite{satheritage}, on which the community is invited to contribute.

The solvers are typically grouped by year of competition.  Images are then named as
\lstinline|satex/<solver-name>:<year>|.

The images are built by compiling solver sources whenever available.  The compiling environment matches with a Linux
distribution of the time of the competition.  We selected the Debian GNU/Linux distribution which provides \docker images
for each of its version since 2000.  For instance, the solvers from the 2000 competition are built using the Debian
``Potato'' as it was back at that time.  In principle, each solver can have its own recipe and environment for building
and execution.  Nevertheless, we managed to devise \docker recipes compatible with several generations of competitions.  The
architecture of the repository also allows custom sets of solvers. For example, the \tool collection includes
the different Knuth's solvers or solvers with Java or Python.

The image building \docker recipes indicate where to download the sources or the binaries whenever the former are not available.
At the time of the writing of this article, most recipes use URL from the website of the SAT competitions.  In order to
provide as most as persistent locations as possible, we are regularly moving more resources on
\zenodo services to host
sources and binaries in a near future \cite{satheritage-zenodo}
(currently, only the
binaries of the original \texttt{satex} and the 2002's competition are hosted on it).

The images can be built locally from the git repository, and are also available for download from the main public \docker
repository \cite{satheritage-docker},
that distributes ``official'' binaries of solvers. This allows
to directly run any collected (or compiled) solver very quickly.

\subsection{Running solvers}

We provide a Python script, called  \lstinline|satex|, which eases the execution and management of available \docker images,
although images can be directly run without it.
The script can be installed using \texttt{pip} utility:  \lstinline|pip3 install -U satex|


The list of available solvers can be fetched
using the command \lstinline|satex list|.

We provide a generic wrapper in each image giving a unified mean to invoke the solver:
a DIMACS file (possibly gzipped) as first argument, and optionally an output file for the
proof:
\begin{lstlisting}
# run a solver on a cnf file
satex run cadical:2019 file.cnf
# run and produce a proof
satex run glucose:2019 file.cnf proof
\end{lstlisting}

The \lstinline|satex info| command gives, together with general information on the solver and the image
environment,
the specific options used for the run.
Alternatively, custom options can be used with the \lstinline|satex run-raw| command.
If the image has not been built locally, it will attempt to fetch it from the online \docker repository.
See the \lstinline|satex -h| for other available commands, such as extracting binaries out of \docker
images and invoking shells within a given image.

\subsection{Building and adding new solvers}

The building of images, which involve the compilation of the solvers when possible, also relies on
\docker images, and thus only requires \docker and Python for the \lstinline|satex| command.
 The following command, executed at the root of the \texttt{sat-heritage/docker-images} repository, will build the matching solvers with their
adequate recipe:
\begin{lstlisting}
satex build '*:2000' # build all 2000 solvers
\end{lstlisting}

Sets of solvers are added by specifying which \docker recipes to use for building the images
and how to invoke the individual solvers. Managing sets of solvers allows
sharing common configurations (such as linux distributions, compilers
and so on) for docker images.
A complete and up-to-date documentation can be found in the README file of the repository.

\section{Ensuring Reproducibility}

Reproducibility is a corner stone of science.
In computer science, it recently appealed for significant efforts by researchers,
institutions and companies to devise good practices and provide adequate infrastructures.
Among the numerous initiatives, Software Heritage \cite{heritage,heritage2} and \zenodo~\cite{zenodo,zenodo2} are
probably the most important efforts for archiving source code, repositories, datasets, and binaries, for which they
provide persistent storage, URLs, and references (DOI).  Another example is the \github Archive Program, a repository on
a 500-years lifespan storage preserved in the Artic World Archive~\cite{AWA}.  Created more recently, the \guix
\cite{guix} initiative aims at keeping the details of any Linux machine configuration, thanks to a declarative system
configuration. External URL used for building any image are also archived. Our tool produces \docker images that can be
easily frozen thanks to \guix, by building \guix images from the Dockerfile recipe. It is also worth mentioning that
\guix has strong connections with Software Heritage and \github.

If we look at reproducibility of SAT solvers experiments on a longer time scale, we can expect that, some day, current
binaries (for i386) will not genuinely run on computers any more. We can expect, however, that there will be i386
emulators. Once such an emulator is set up, we can also expect \docker to be available on it, and then all the
images we built will be handled natively. If not, as \docker recipes are plain text, it will be easy to convert them to another framework.

Therefore, facilitating the accessibility of software in time now boils down to simple habits, such as using source
versioning platforms, taking advantage of services like \zenodo or Software Heritage to freeze packages dependencies,
source code, binaries, and benchmarks, and provide \docker images to give both environments and recipes to build and run
your software.

\section{Conclusion}

We presented a tool for easily archiving and running all SAT solvers produced so far. Such a tool is needed because
of (1) source code and experiments are crucial for the SAT community and (2) there are already too many SAT solvers
produced so far, with many different ways of publishing sources. 

In order to complete our tool we think at further improvements, like including \docker images for compiling SAT solvers
for other architectures than i386 (ARM for instance), but also initiating another important effort for the community:
including \docker images for benchmarks generations and maintenance. Many benchmarks are combinatoric ones, typically
generated by short programs. These generators are generally not distributed by the different competitive events and
may contain a lot of information on the structure of the generated problems. We also think that our tool could be very
interesting for SAT solvers configurations and easy cloud-deployment in a portfolio way.
We also expect our work to give the community the best possible habits for state of the art archiving and reproducibility
practices.

\bibliography{biblio}
\bibliographystyle{plain}
\end{document}